\documentclass[11pt]{article}

\usepackage[preprint]{acl}

\usepackage{times}
\usepackage{latexsym}
\usepackage{amsmath}
\usepackage{subcaption}
\usepackage{booktabs} 
\usepackage{makecell}

\usepackage[T1]{fontenc}

\usepackage[utf8]{inputenc}

\usepackage{microtype}

\usepackage{inconsolata}

\usepackage{graphicx}

\title{Confidence-Adaptive SwiGLU for Mixture-of-Experts}

\author{{\bf Shaohua Li$^1$} \hspace{1em} {\bf Xiuchao Sui$^1$} \hspace{1em} {\bf Xiaobing Sun$^1$} \hspace{1em} {\bf Yuhang Wu$^2$} \\
         {\bf Liangli Zhen$^1$} \hspace{1em} {\bf Yong Liu$^1$} \hspace{1em} {\bf Rick Siow Mong Goh$^1$} \\
         $^1$Institute of High Performance Computing, Agency for Science, Technology and Research, Singapore \\
         $^2$Shanghai University of Engineering Science, China}

\begin{document}
\maketitle
\begin{abstract}

SwiGLU has become a standard gated activation in modern Transformer MLPs, yet its gate sharpness---the smoothness and selectivity of the gating function---is typically fixed throughout training. In this work, we propose Confidence-Aware SwiGLU ($\kappa$-SwiGLU), a variant of SwiGLU for Mixture-of-Experts (MoE) models that adjusts expert gate sharpness according to token-level routing confidence. Specifically, $\kappa$-SwiGLU parameterizes the SiLU gate sharpness coefficient as a learnable function of the router logit, enabling each expert gate unit to interpolate between smooth, broadly active gating and sharp, selective gating. We evaluate $\kappa$-SwiGLU on the FineWeb-Edu dataset across MoE Transformer models ranging from 8 to 28 layers. Across these settings, $\kappa$-SwiGLU improves mean CORE performance while adding negligible parameters and incurring only a small computational overhead, demonstrating that confidence-aware gate sharpness is a promising mechanism for improving MoE MLPs. The code is available at \url{https://github.com/askerlee/kappa-swiglu}.

\end{abstract}

\section{Introduction}

\begin{figure}[t]
  \centering
  \includegraphics[width=\columnwidth]{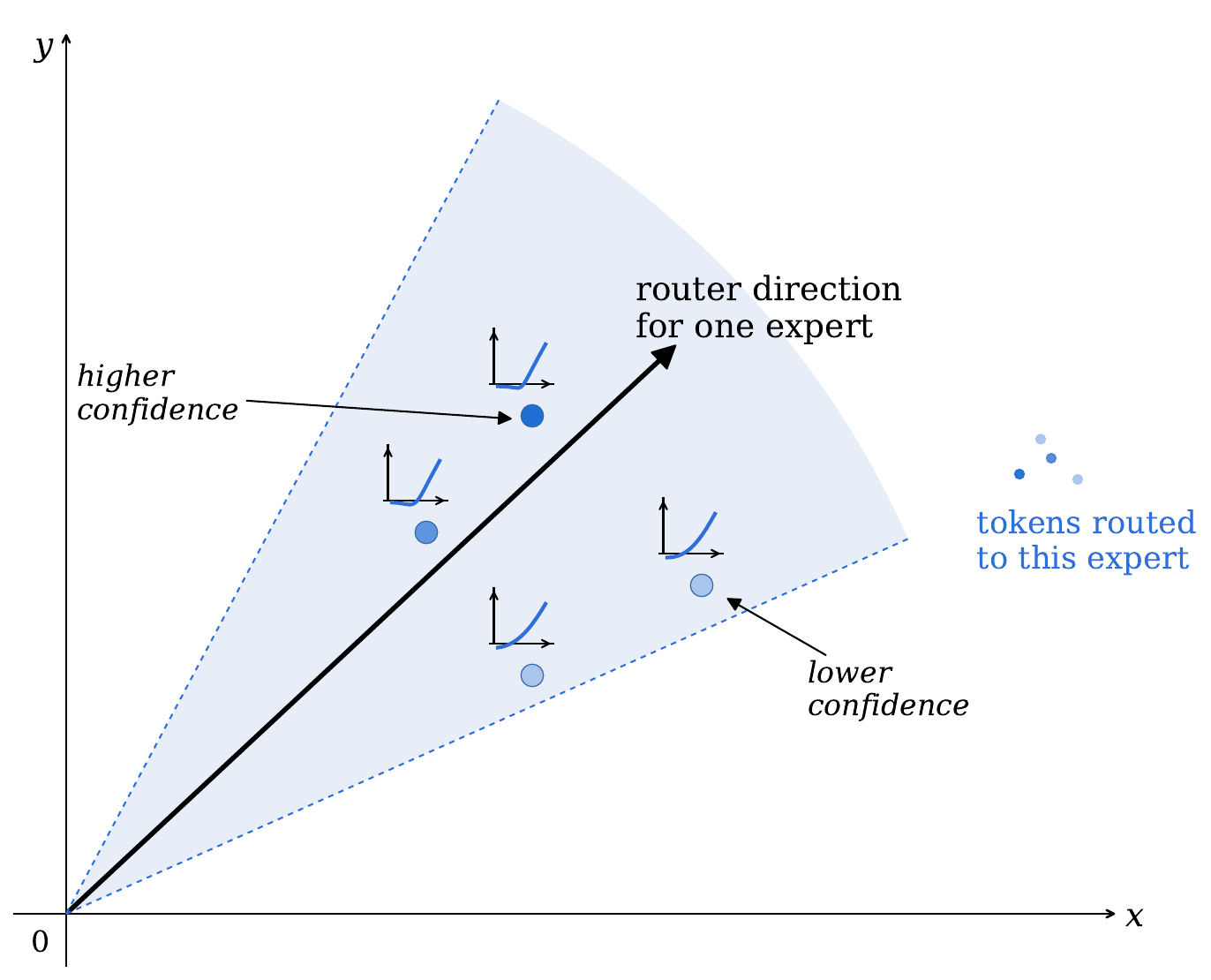}
\caption{Illustration of how routing confidence modulates gate sharpness in a $\kappa$-SwiGLU instance. The shaded region denotes the subset of embedding space routed to a particular expert. Tokens closer to the expert's router weight vector (center arrow) have higher routing confidence, which can induce different gate sharpness depending on the learned confidence--sharpness mapping. The small SiLU curves illustrate the corresponding gate-sharpness changes; Figure~\ref{fig:bias-sharpness-contrast} provides a magnified view of this mechanism.}
\label{fig:k-swiglu}
\end{figure}
SwiGLU MLPs \cite{gated-conv,glu} have become a standard component of modern Transformer architectures, including dense language models \cite{llama2,qwen3} and Mixture-of-Experts (MoE) models \cite{deepseek-moe,mixtralexperts,qwen3,nemotron3,gpt-oss,glm45,kimik2}. 
In a SwiGLU MLP, a SiLU gate modulates intermediate activations conditioned on the input, selectively suppressing or amplifying features and improving expressivity at low computational cost.

The SwiGLU layer is commonly defined as:
\begin{align*}
\mathrm{SwiGLU}(x)
&= \mathrm{SiLU}(W_g x) \odot (W_u x), \\
\mathrm{SiLU}(z)
&= z \cdot \sigma(z) = \frac{z}{1 + e^{-z}} .
\end{align*}
SiLU can be viewed as a fixed-sharpness instance of Swish. We denote the corresponding sharpness-adjusted SiLU gate as
\begin{align*}
\mathrm{SiLU}_{\kappa}(z)
&= z \cdot \sigma(\kappa z),
\end{align*}
where $\kappa$ is a sharpness coefficient controlling the transition between inactive and active gate states. The standard SiLU gate corresponds to $\kappa=1$\footnote{This sharpness-adjusted SiLU gate is equivalent to the Swish activation, with $\kappa$ corresponding to the Swish sharpness coefficient. Although Swish allows this coefficient to be learned in principle \cite{swish}, modern Transformer MLPs typically use the fixed SiLU gate of $\kappa=1$.}. 

\begin{figure*}[t]
  \centering
  \includegraphics[width=1.8\columnwidth,trim=0 2cm 0 0,clip]{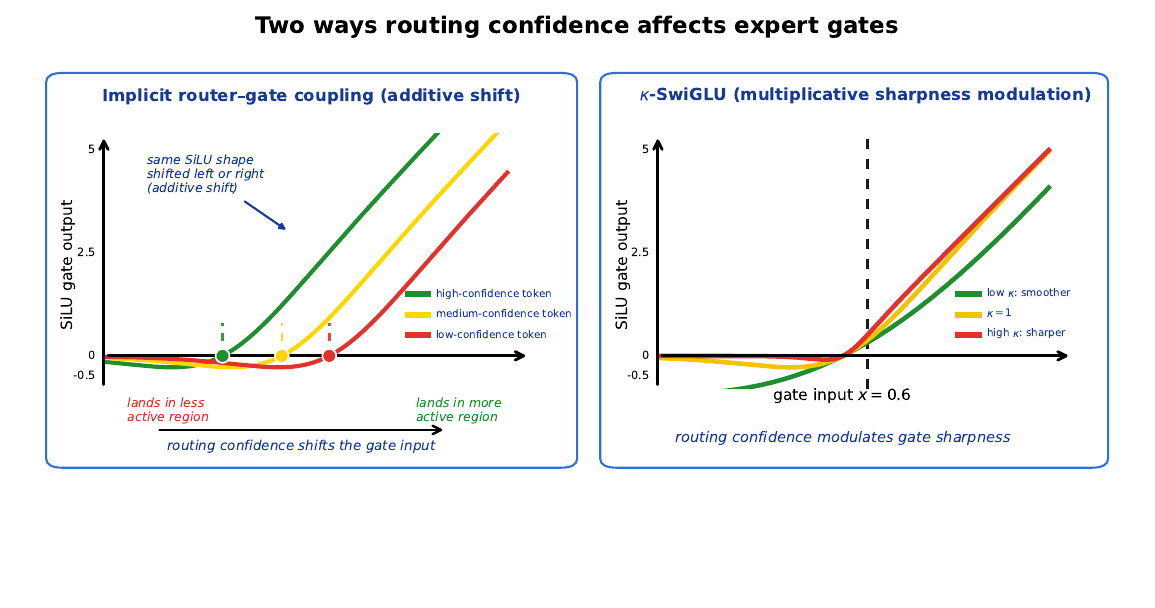}
  \caption{Two mechanisms by which routing confidence can influence expert gates. Left: naturally emerging router--gate alignment implicitly shifts the input to the SiLU gate, moving tokens across different regions of the same gate curve. Right: $\kappa$-SwiGLU explicitly changes the gate curve itself by modulating its sharpness with a token-dependent multiplicative coefficient.}  
\label{fig:bias-sharpness-contrast}
\end{figure*}

Larger $\kappa$ yields a sharper, more selective gate, while smaller $\kappa$ produces smoother, more broadly active gating. Allowing this sharpness to adapt could provide a more expressive mechanism for regulating feature activation. This is especially relevant in MoE models, where the router dynamically assigns each token to a small subset of experts based on router logits. These scores provide a natural signal of routing certainty: when a token receives a high score for an expert, the router is more confident in that token--expert assignment; when the score is lower, the assignment is more uncertain. Beyond expert selection, such scores could therefore serve as a control signal for modulating gate activations inside the selected experts.

We identify a previously overlooked factor in training MoEs with SwiGLU experts: a hidden co-evolution between the router and the expert gate. Specifically, we observe that gate projection directions within an expert rapidly become aligned or anti-aligned with the corresponding router weight vector during training. This alignment makes the expert gate implicitly sensitive to routing certainty: tokens with different router affinities are shifted to different regions of the SiLU gate's transition curve, causing subsets of expert activations to be systematically amplified or suppressed, as illustrated in the left panel of Figure~\ref{fig:bias-sharpness-contrast}. This emergent coupling further motivates using routing confidence to regulate how each selected expert processes a token.

Motivated by this implicit router--gate coupling, we propose $\kappa$-SwiGLU, which explicitly uses routing confidence to modulate the sharpness of each expert's SiLU gate. Unlike the emergent alignment effect, which influences gate behavior by \emph{additively shifting} the gate input, $\kappa$-SwiGLU directly adjusts the gate's smoothness and selectivity through a token-dependent \emph{multiplicative sharpness coefficient}. Each expert learns its own confidence--sharpness mapping, allowing routing confidence to induce either sharper, more selective gates or smoother, more broadly active gates depending on the learned parameters. This mechanism is illustrated in Figure~\ref{fig:k-swiglu} and in the right panel of Figure~\ref{fig:bias-sharpness-contrast}, where it is contrasted with the implicit additive-shift effect shown in the left panel.

We evaluate $\kappa$-SwiGLU by training SwiGLU-based MoE models on FineWeb-Edu, scaling from 8 to 28 layers. Across model settings, $\kappa$-SwiGLU shows a consistent positive trend in model quality, as reflected in stronger pretraining benchmark performance. Our analysis suggests that confidence-aware gate sharpness enables more expressive gating patterns and promotes synergistic interactions between the router and expert gates.

\section{Related Work}
\vspace{-5pt}
\paragraph{MoE Load balancing and routing stability.}
One of the central challenges in MoE training is preventing routing collapse, where only a small subset of experts receives most tokens. Prior work commonly addresses this issue using auxiliary load-balancing losses that encourage more uniform expert utilization \cite{outrageous,gshard,switch-transformer,stablemoe,st-moe}. Switch Transformer \cite{switch-transformer} also introduced the router $z$-loss to regularize router logits. Subsequent work has studied how load-balancing losses are sensitive to implementation details such as batch statistics \cite{loss-details}, proposed auxiliary-loss-free balancing strategies \cite{auxiliary-free}, and alternative balancing formulations such as $\phi$-balancing \cite{phi-balancing}. These approaches primarily regulate how tokens are assigned to experts. In contrast, $\kappa$-SwiGLU uses the selected expert's router signal to modulate gate sharpness inside the expert MLP.\vspace{-5pt}
\paragraph{Geometry of MoE routing.}
Recent work has explored geometry-aware routing mechanisms, including Routing Manifold Alignment and kNN-augmented routing, which encourage expert assignment to better reflect token-representation geometry \cite{routing-manifold,routing-knn}. Other studies address representational collapse in MoE training \cite{simsmoe}, geometric regularization of expert weights and activations \cite{expert-dissimilarity}, and auxiliary losses that couple routers and experts \cite{coupling-loss}. Rather than adding geometric constraints, our work uses the router logit directly as a token-level signal for adaptive gate sharpness.\vspace{-5pt}
\paragraph{Gated activations in Transformer MLPs.}
Early Transformer models commonly used ReLU- or GELU-based feed-forward networks \cite{relu,gelu}, while modern LLMs have largely converged toward gated variants such as GLU, GeGLU, and SwiGLU \cite{gated-conv,glu}. SwiGLU has become a common choice due to its strong empirical performance and modest computational cost. Recent work has revisited this activation design space, including ReLU$^2$ \cite{relu2}, expanded gating ranges in xGELU and xSiLU \cite{xsilu}, and adaptive mixtures of activation functions \cite{mix-of-activations}. In practice, SwiGLU gates can require stabilization techniques such as clamping to avoid excessively large activations during training \cite{gpt-oss,deepseekv4}; relatedly, Power Linear Unit (PowLU) controls activation magnitude using a bounded rational power function \cite{powlu}. However, the interaction between expert routing and gated activations remains relatively underexplored. Our work studies this interaction by using MoE routing confidence to adapt the sharpness of the SwiGLU gate.

\section{Method}

In MoEs, routed tokens tend to be biased toward the corresponding router direction. Meanwhile, expert gate projection vectors naturally become aligned or anti-aligned with the same router direction during training. These two effects induce an implicit, confidence-induced gate bias in the expert gates: tokens with different router logits are shifted to different regions of the SiLU gate. 

To complement this implicit bias mechanism, we propose $\kappa$-SwiGLU, which uses router logits to produce a token-dependent sharpness coefficient for each expert gate. Unlike the emergent implicit gate bias which shifts the gate input additively, $\kappa$-SwiGLU modulates the gate multiplicatively, allowing each expert to adapt its activation selectivity according to routing confidence.

\subsection{Router Logits as Routing Confidence}
\begin{figure}[t]
  \centering
  \includegraphics[width=0.8\columnwidth]{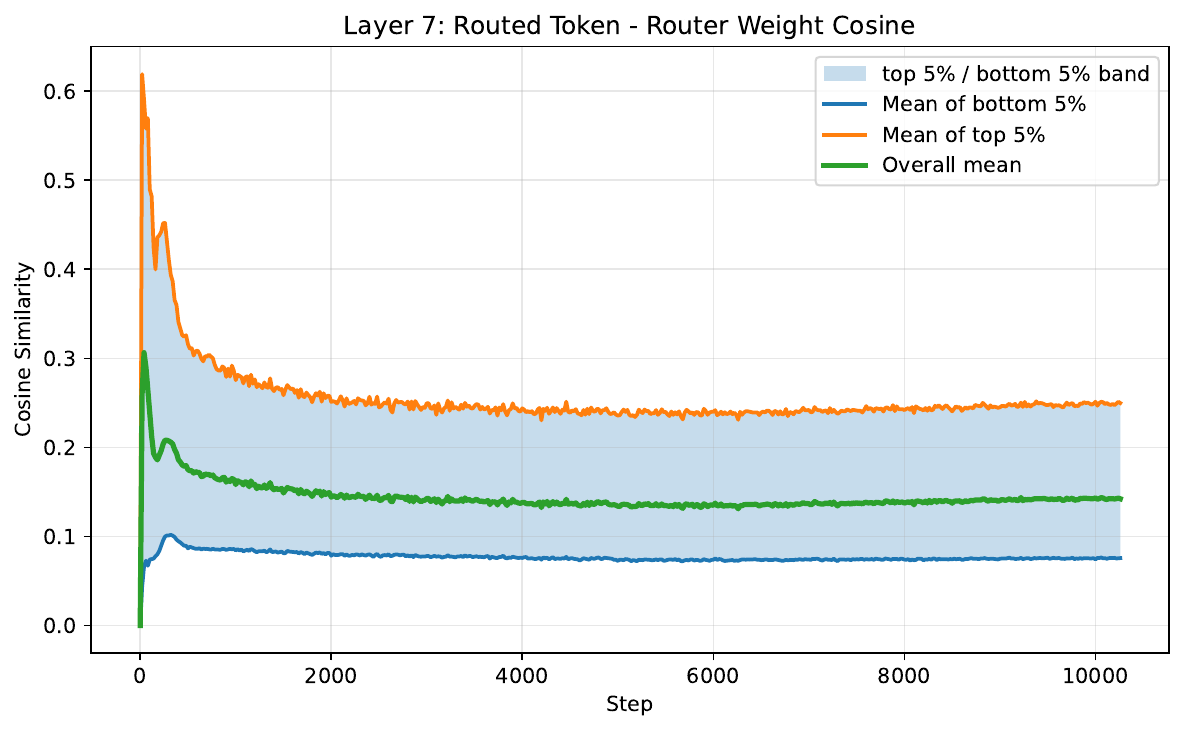}
\caption{Cosine similarity between router weight vectors and routed token embeddings during training, measured over all routed token--expert pairs in the 7th layer of an 8-layer MoE. The similarities stabilize between $0.075$ and $0.25$, with a mean of $0.15$. It indicates that tokens routed to an expert have non-negligible alignment with its router direction, yielding high router logits.}
\label{fig:router-token-similarity}
\end{figure}

Mixture-of-Experts (MoE) models typically route each token to a small subset of experts according to router logits, computed from the inner product between an expert router vector $r_e$ and the input token representation $x$. Geometrically, the tokens routed to expert $e$ occupy a narrow region of representation space biased toward the router direction $r_e$\footnote{For a random unit vector in 512-D, the volume fraction satisfying $\cos(x,r_e)\ge 0.15$ is approximately $0.03\%$.}, as illustrated in Figure~\ref{fig:router-token-similarity}. Tokens that are more closely aligned with $r_e$ receive higher router scores and can therefore be interpreted as higher-confidence assignments to expert $e$. Based on this observation, we use the \emph{router logit} as a token-level routing confidence signal\footnote{We do not use the routing probability because it depends on the set of competing experts selected for the token and on the normalization over their logits, which introduces additional variation unrelated to the token--expert affinity itself.}
\begin{align}
s_e(x) = r_e^\top x \label{router-logits}
\end{align}

\vspace{-5pt}
\begin{figure*}[th]
  \centering
  \begin{subfigure}[t]{0.4\linewidth}
    \centering
    \includegraphics[width=\linewidth]{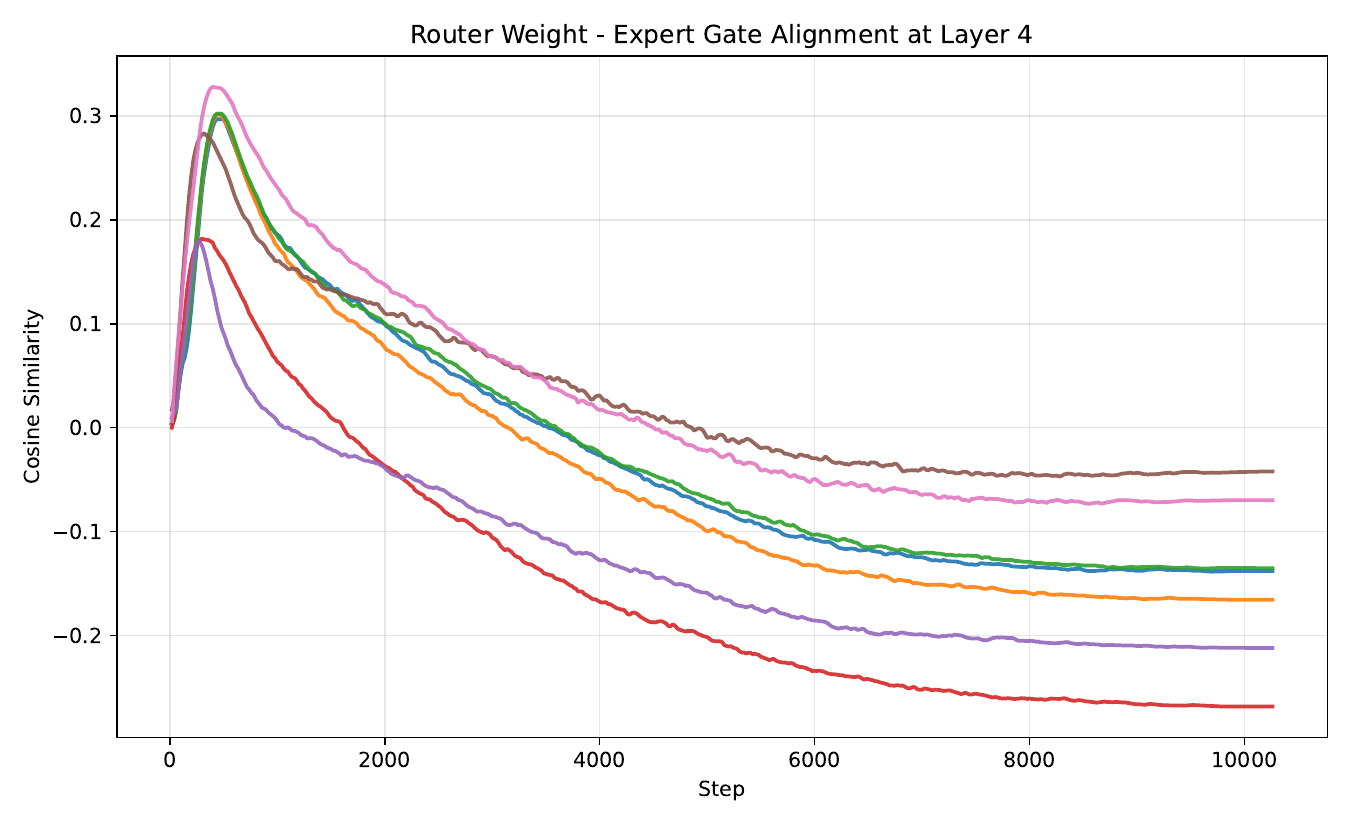}
    \caption{Layer 4}
    \label{fig:router-gate-align-early}
  \end{subfigure}
  \hfill
  \begin{subfigure}[t]{0.4\linewidth}
    \centering
    \includegraphics[width=\linewidth]{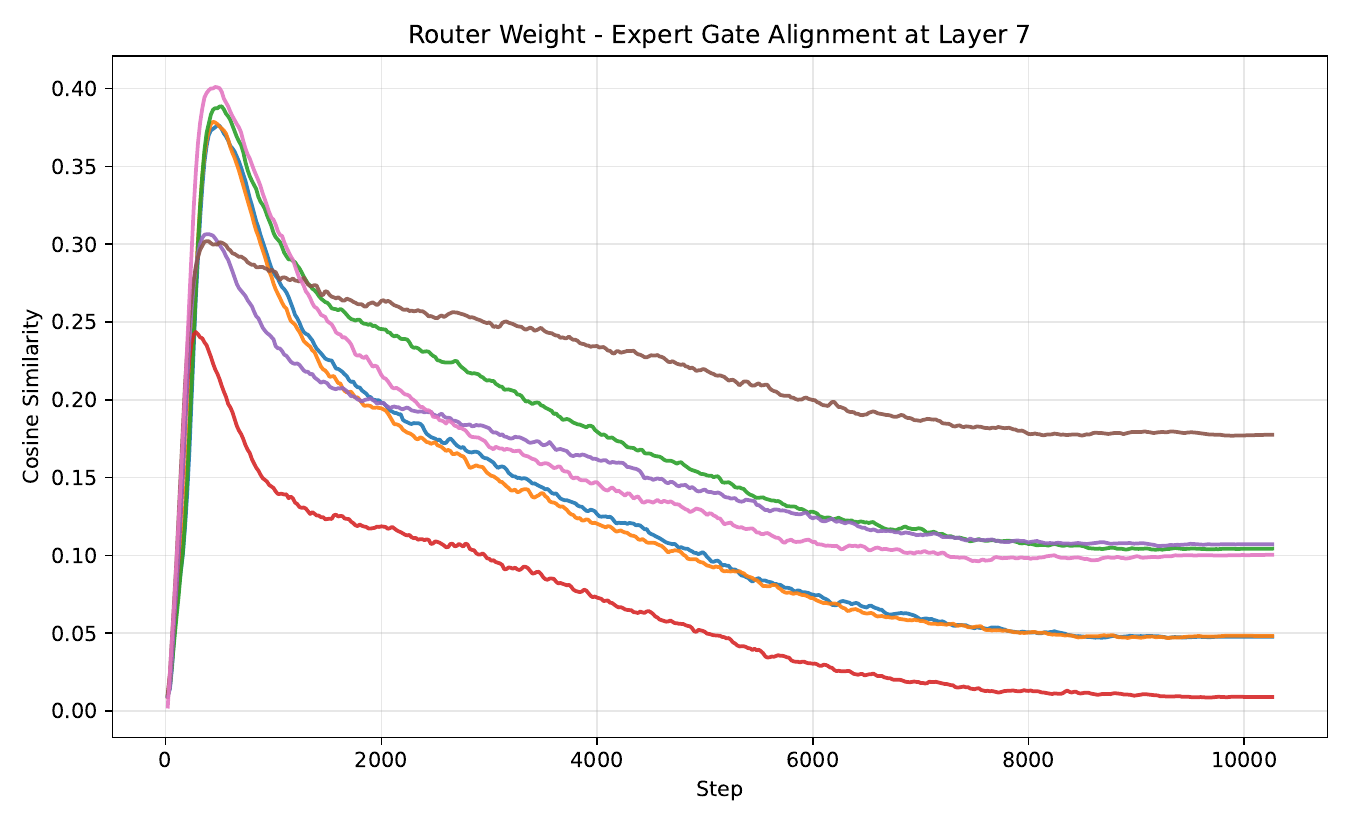}
    \caption{Layer 7}
    \label{fig:router-gate-align-late}
  \end{subfigure}
\caption{Router--gate alignment over training for two representative layers. We report the average cosine similarity between each router weight vector and the corresponding expert's gate projection vectors across 7 independently trained 8-layer MoE models. Layer 4 rapidly develops positive router--gate alignment within the first few hundred steps, but later becomes consistently negative across runs. Layer 7 maintains positive alignment for most of training, although its magnitude also decays over time. This suggests that router--gate coupling emerges broadly during training while exhibiting layer-dependent signed dynamics.}
\label{fig:router-gate-alignment}
\end{figure*}

\subsection{Confidence-Induced Gate Bias}

We first empirically verify the existence of router--gate alignment by tracking the cosine similarity between each expert's router vector and its gate projection vectors during training. Figure~\ref{fig:router-gate-alignment} shows the router--gate alignment dynamics for two representative layers across 7 independently trained 8-layer MoE models. Expert gate projections rapidly become aligned with the corresponding router vector within the first few hundred training steps, reaching peak cosine similarities of $0.2$--$0.4$. Although the coupling strength changes over training and varies across layers, it remains non-negligible throughout. This suggests that router--gate coupling emerges naturally during training while exhibiting diverse, layer-dependent signed dynamics.

To better understand the impact of router--gate alignment, let $w_{e,j}$ denote the $j$-th gate projection vector of expert $e$, and let $\hat r_e = r_e / \|r_e\|_2$ be the unit-normalized router vector. We decompose $w_{e,j}$ into components parallel and orthogonal to the router direction:
\[
w_{e,j} = (w_{e,j}^{\top}\hat r_e)\hat r_e + w_{e,j}^{\perp}.
\]
Similarly, we write the input representation as $x = a\hat r_e + x^\perp$, where $a = \hat r_e^\top x$. The gate input can then be written as
\[
w_{e,j}^{\top}x
=
(w_{e,j}^{\top}\hat r_e)a
+
(w_{e,j}^{\perp})^\top x^\perp.
\]
The first term, $(w_{e,j}^{\top}\hat r_e)a$, acts as a \emph{confidence-induced gate bias}: tokens with larger projection onto the router direction induce a systematic shift in the corresponding gate input. Thus, router--gate alignment implicitly changes the bias of expert gates, as illustrated in the left panel of Figure~\ref{fig:bias-sharpness-contrast}. 

\begin{figure}[t]
  \centering
  \includegraphics[width=0.8\columnwidth]{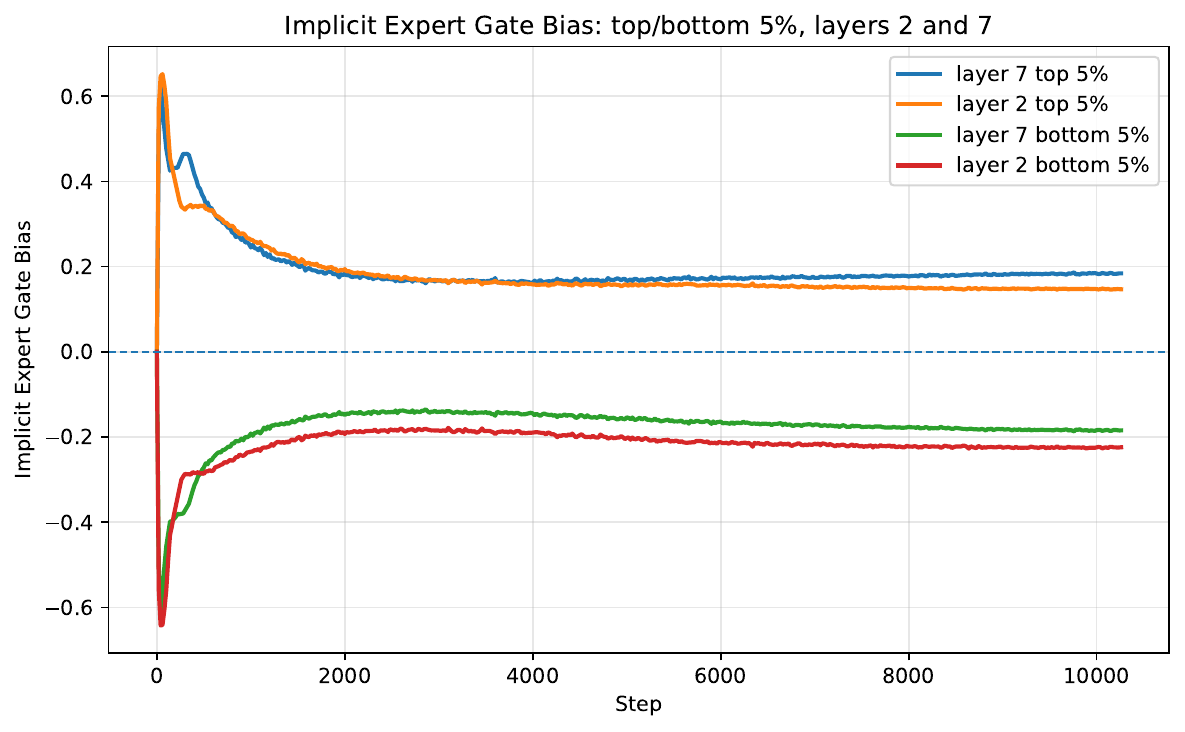}
  \caption{Empirically observed implicit bias in expert gates induced by router--gate alignment. We compute the bias term $(w_{e,j}^{\top}\hat r_e)a$ for each gate projection across all experts in two representative layers of an 8-layer MoE, and report the mean of the top and bottom $5\%$ values. The top $5\%$ biases remain consistently positive, while the bottom $5\%$ remain consistently negative, indicating that router--gate alignment can systematically shift subsets of expert gates in opposite directions.}\label{fig:implicit_bias}
\end{figure}

Figure~\ref{fig:implicit_bias} empirically confirms this effect: the top and bottom $5\%$ of induced bias values remain substantially positive and negative throughout training, indicating that router--gate alignment produces non-negligible bidirectional shifts in expert gates.

\subsection{Confidence-Adaptive SwiGLU}
Motivated by the naturally emerging confidence-induced gate bias, we introduce an explicit confidence-aware sharpness parameterization to make the router--gate coupling more expressive and controllable. While the emergent alignment effect modulates expert gates implicitly through confidence-induced gate bias shifts, $\kappa$-SwiGLU directly controls the smoothness and selectivity of the SiLU gate using a token-dependent sharpness coefficient. This provides a more flexible mechanism for routing confidence to influence expert computation, allowing each expert to learn whether higher confidence should induce sharper, more selective gating or smoother, more broadly active gating.

We first define a sharpness-adjusted SiLU gate:
\begin{align}
\mathrm{SiLU}_{\kappa}(z)
&= z \cdot \sigma(\kappa z),
\end{align}
where $\kappa$ is a positive sharpness coefficient. 

\begin{figure}[t]
  \centering
  \includegraphics[width=1\columnwidth]{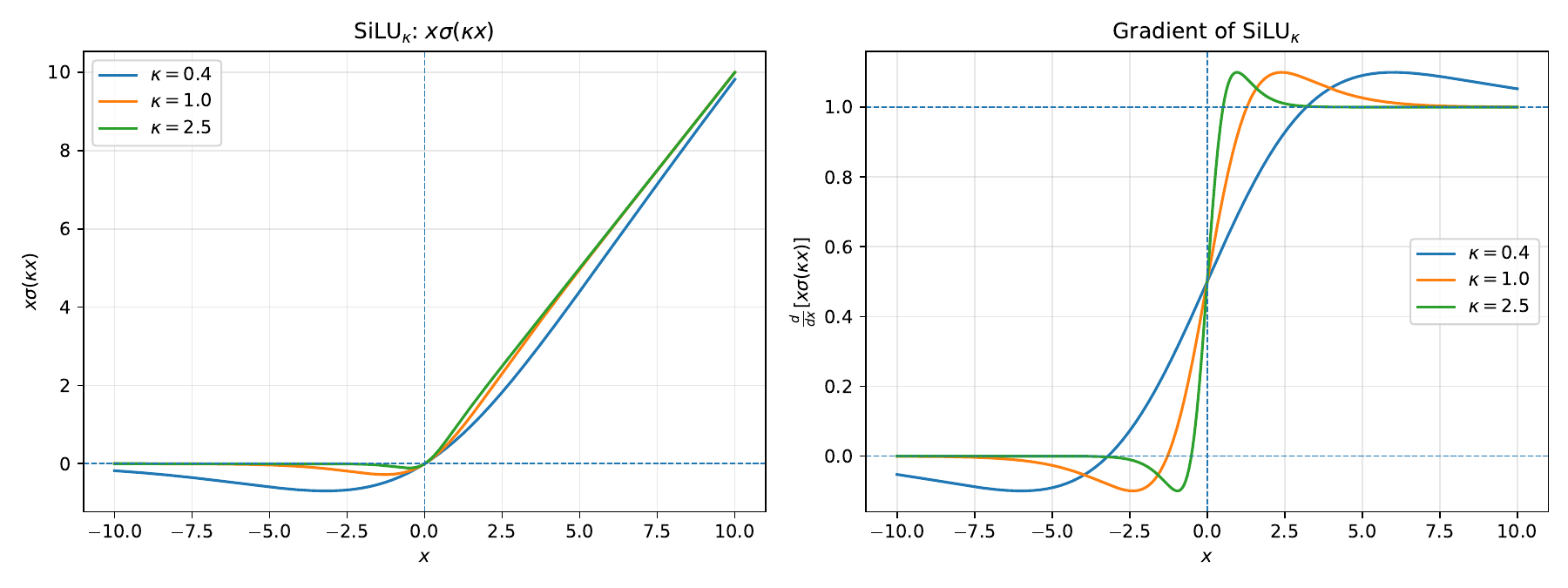}
\caption{The $\mathrm{SiLU}_{\kappa}(z)$ function under different sharpness coefficients $\kappa$. Larger $\kappa$ yields a sharper transition between inactive and active states around zero, while smaller $\kappa$ yields a smoother transition. The right panel shows the corresponding gradient, $\frac{d}{dz}\mathrm{SiLU}_{\kappa}(z)$, where different $\kappa$ values lead to substantially different gradient profiles near the transition region around zero.}
\label{fig:silu-kappa}
\end{figure}

The standard SiLU gate corresponds to the fixed-sharpness case $\kappa=1$. Larger values of $\kappa$ produce sharper and more selective gating, while smaller values produce smoother and more broadly active gating. To intuitively understand the effect of $\kappa$, we plot $\mathrm{SiLU}_{\kappa}(z)$ and its gradient under different $\kappa$ values in the left and right panels of Figure~\ref{fig:silu-kappa}, respectively. As $\kappa$ increases, the transition between inactive and active states becomes sharper, with a more pronounced change around zero. Conversely, as $\kappa$ decreases, the transition becomes smoother, with a more gradual change across a wider range of inputs. These differences are most pronounced in the transition region around zero, where the gate output changes rapidly with respect to the input, while different $\kappa$ values produce more similar behavior in the saturated regions far from zero. In the gradient panel of Figure~\ref{fig:silu-kappa}, the corresponding gradients show even greater sensitivity to $\kappa$: larger $\kappa$ yields steeper gradient changes around the transition region, indicating that the gate becomes more responsive to small variations in near-zero inputs.

We then parameterize the sharpness coefficient $\kappa$ as a learnable function of the routing confidence:
\begin{align}
\kappa_{e,j}(x) &= \phi\left(\alpha_{e,j} \cdot s_e(x) + b_{e,j}\right),
\label{eq:kappa}
\end{align}
where $s_e(x)$ is the router logit defined in Eq.~\eqref{router-logits}, $\alpha_{e,j}$ and $b_{e,j}$ are learnable parameters, and $\phi$ is a monotonically increasing function that maps the potentially unbounded confidence-conditioned signal to a positive sharpness coefficient.

To avoid extremely large or small sharpness values that could destabilize training, we use the following bounded exponential mapping for $\phi$:
\begin{align}
\phi(z) &= U^{\tanh(z)},
\end{align}
where $U > 1$ is a hyperparameter controlling the range of the sharpness coefficient. 

This choice of $\phi$ has two useful properties. First, since $\tanh(z) \in (-1,1)$, it maps the confidence-conditioned signal to a bounded sharpness coefficient, $\kappa_{e,j}(x) \in (1/U, U)$, preventing extreme $\kappa$ values. Second, when $z=0$, we have $\tanh(z)=0$ and $\phi(z)=1$, so $\kappa$-SwiGLU reduces to the standard SwiGLU gate.

Compared with the original SwiGLU,
\begin{align}
&\quad\quad \mathrm{SwiGLU}_e(x)
=
\mathrm{SiLU}(W_{g,e}x) \odot (W_{u,e}x) \nonumber \\
&=
\left[
(W_{g,e}x)
\odot
\underbrace{
\sigma\left(W_{g,e}x\right)
}_{\text{fixed sharpness}}
\right]
\odot
(W_{u,e}x),
\end{align}
$\kappa$-SwiGLU replaces the fixed-sharpness SiLU gate with the confidence-conditioned gate $\mathrm{SiLU}_{\kappa_e(x)}$:
\begin{align}
& \mathrm{\kappa\text{-}SwiGLU}_e(x)
=
\mathrm{SiLU}_{\kappa_e(x)}(W_{g,e}x) \odot (W_{u,e}x) \nonumber \\
&=
\left[
(W_{g,e}x)
\odot
\underbrace{
\sigma\left(\kappa_e(x) W_{g,e}x\right)
}_{\text{confidence-aware sharpness}}
\right] \odot (W_{u,e}x).
\end{align}
For notational simplicity, we omit the gate-unit index $j$ here. Thus, $\kappa$-SwiGLU generalizes the fixed-sharpness gate in standard SwiGLU by allowing each expert to adapt its activation selectivity according to token-level routing confidence. When $\kappa_{e,j}(x)>1$, the gate becomes sharper and more selective than the standard SiLU gate; when $\kappa_{e,j}(x)<1$, it becomes smoother and more broadly active, as illustrated in Figure~\ref{fig:silu-kappa}. 

\subsection{Regularization on $\kappa$ parameters}
We apply L2 regularization to $\alpha_{e,j}$ and $b_{e,j}$ to prevent the sharpness modulation from deviating too aggressively from the standard SiLU:
\begin{align}
\mathcal{L}_{\mathrm{reg}}
=
\lambda_{\alpha} \sum_{e,j} \alpha_{e,j}^2
+
\lambda_b \sum_{e,j} b_{e,j}^2.
\end{align}

\section{Experiments}

\begin{table*}[t]
\centering
\small
\begin{tabular}{lccccccccc}
\toprule
Model & Layers & \makecell{MoE\\Layers} & \makecell{Dense\\Layers} & \#Experts & Top-$k$ & \makecell{Total Params\\(M)} & \makecell{Active Params\\(M)} & \makecell{Tokens\\(B)} & \makecell{GPU\\Hours} \\
\midrule
MoE-8L       & 8  & 6  & 2  & 64  & 2 & 2,905 & 269   & 2.7  & 6.3 \\
MoE-10L      & 10 & 8  & 2  & 32  & 2 & 3,526 & 504   & 4.5  & 10.0 \\
MoE-12L      & 12 & 10 & 2  & 32  & 2 & 4,509 & 685   & 5.9  & 17.3 \\
MoE-14L      & 14 & 10 & 4  & 16  & 2 & 4,430 & 1,098 & 8.0  & 33.3 \\
Sandwich-16L & 16 & 2  & 14 & 128 & 2 & 1,035 & 241   & 4.4  & 5.0 \\
Sandwich-20L & 20 & 2  & 18 & 128 & 2 & 1,633 & 393   & 7.0  & 10.2 \\
Sandwich-24L & 24 & 2  & 22 & 128 & 2 & 2,378 & 593   & 10.3 & 15.7 \\
Sandwich-28L & 28 & 2  & 26 & 128 & 2 & 3,279 & 849   & 14.2 & 17.3 \\
\bottomrule
\end{tabular}
\caption{Model configurations and training budgets. We report the total number of Transformer layers, the number of MoE and dense layers, the number of experts, the MoE routing top-$k$, the total and active parameter counts, the number of training tokens in billions, and the GPU-hour cost for a single training run on H200 GPUs.}
\label{tab:model-configs}
\end{table*}

\subsection{Experimental Details}

We implement $\kappa$-SwiGLU in a standard MoE Transformer architecture with SwiGLU MLPs. Our training pipeline is based on the widely used Nanochat codebase\footnote{\url{https://github.com/karpathy/nanochat/}}, with modifications to incorporate a standard token-choice router and the proposed $\kappa$-SwiGLU activation. We train all models on the FineWeb-Edu dataset \cite{fineweb-edu}. 

For matrix parameters, we use Aurora \cite{aurora}, an emerging state-of-the-art optimizer for large language model training, with a learning rate of $0.01$ and weight decay of $0.05$. For non-matrix parameters, we use AdamW with a learning rate of $0.3$ and betas $(0.8, 0.95)$. These hyperparameters are inherited from the default Nanochat settings for dense models and slightly adapted to better accommodate the MoE architecture. An auxiliary load-balancing loss \cite{switch-transformer} is applied to the router logits to encourage balanced expert utilization, with weight $10^{-3}$. A router $z$-loss \cite{switch-transformer} is also applied to suppress excessively large router logits, with weight $10^{-5}$. 

All models are trained using four H200 GPUs, each equipped with 141GB of memory. To fit within the available GPU memory, we vary the number of candidate experts and number of MoE layers across settings, as detailed in the next subsection. Table~\ref{tab:model-configs} summarizes the model configurations and training budgets for all settings.

We evaluate pretraining performance using the average score across 22 CORE benchmarks \cite{core}, which cover a diverse set of tasks including textbook knowledge, commonsense reasoning, and language modeling. We report Centered CORE accuracy, computed as the average benchmark score relative to a fixed-answer baseline. This metric mitigates potential biases arising from differences in answer format, class imbalance, and task difficulty across benchmarks.

Since models trained with the same settings but different random seeds can show noticeable variation on the CORE benchmark \cite{nanochat_leaderboard}, we train three independent runs for each setting using random seeds 24, 26, and 28. We report the mean and standard deviation across these runs to ensure that our results are robust to random variation.

\subsection{Model Settings and Training Budgets}
To fit within the available GPU memory, we adopt three memory-saving strategies. First, we use at most 10 MoE layers and implement any remaining layers as dense layers. Second, as model depth increases, we gradually reduce the number of candidate experts from 64 to 16, reducing the memory cost of MoE layers at the expense of a smaller expert pool. Third, for deeper models with 16--28 layers, we adopt a sandwiched MoE architecture, in which only the middle two layers are MoE layers and all remaining layers are dense. Despite using only two MoE layers in this setting, $\kappa$-SwiGLU yields gains in the deeper sandwiched models, suggesting that confidence-aware gate sharpness can also benefit mixed MoE-dense architectures.

As summarized in Table~\ref{tab:model-configs}, we train standard MoE models with 8, 10, 12, and 14 layers, consisting of 2--4 dense layers and 6--10 MoE layers, as well as sandwiched MoE models with 16, 20, 24, and 28 layers, consisting of 14--26 dense layers and 2 MoE layers. We set the MoE routing top-$k$ to $2$ for all models. 

Since MoE models contain substantially more total parameters than dense models, we use a token-to-parameter ratio of 5 for all models. Following the common observation that not all MoE parameters are activated for each token, we estimate the effective parameter count using a square-root scaling rule, detailed in the appendix~\ref{app:scaling}.

\subsection{Optimization of $\kappa$ Parameters}

Optimizing $\kappa$-SwiGLU requires learning the gate-wise parameters $\alpha_{e,j}$ and $b_{e,j}$ for each expert. Since $\alpha_{e,j}$ and $b_{e,j}$ are scalars, they introduce only a negligible increase in parameter count compared with the original SwiGLU. Moreover, the router logits $s_e(x)$ are already available during the forward pass, so computing $\kappa_{e,j}(x)$ requires no additional matrix multiplications and only a few elementwise operations, resulting in a small computational overhead.

The range hyperparameter $U$ is set to $3$, constraining $\kappa$ to the interval $(1/3, 3)$. For L2 regularization of $\alpha_{e,j}$ and $b_{e,j}$, we set $\lambda_{\alpha}=2\times10^{-2}$ and $\lambda_b=10^{-2}$, which we find effective in preventing overfitting while preserving sufficient flexibility for learning sharpness modulation.

To ensure stable training, we initialize $\alpha_{e,j}=b_{e,j}=0$, so that $\phi(\alpha_{e,j}s_e(x)+b_{e,j})=\phi(0)=1$ for all tokens. This initializes $\kappa$-SwiGLU to the standard SiLU gate. During the first $1/10$ of training iterations, we keep $\alpha_{e,j}$ and $b_{e,j}$ frozen at 0, allowing the model to establish stable initial routing behavior and expert representations before introducing confidence-aware sharpness modulation. After this initial phase, we unfreeze $\alpha_{e,j}$ and $b_{e,j}$ and update them by backpropagation together with the rest of the model parameters. We use a learning rate schedule of linear warmup followed by linear decay: the learning rate is warmed up to $0.12$ during the first 1000 iterations and then linearly decayed to $0.06$ by the end of training.

\begin{figure}[t]
  \centering
  \includegraphics[width=\columnwidth]{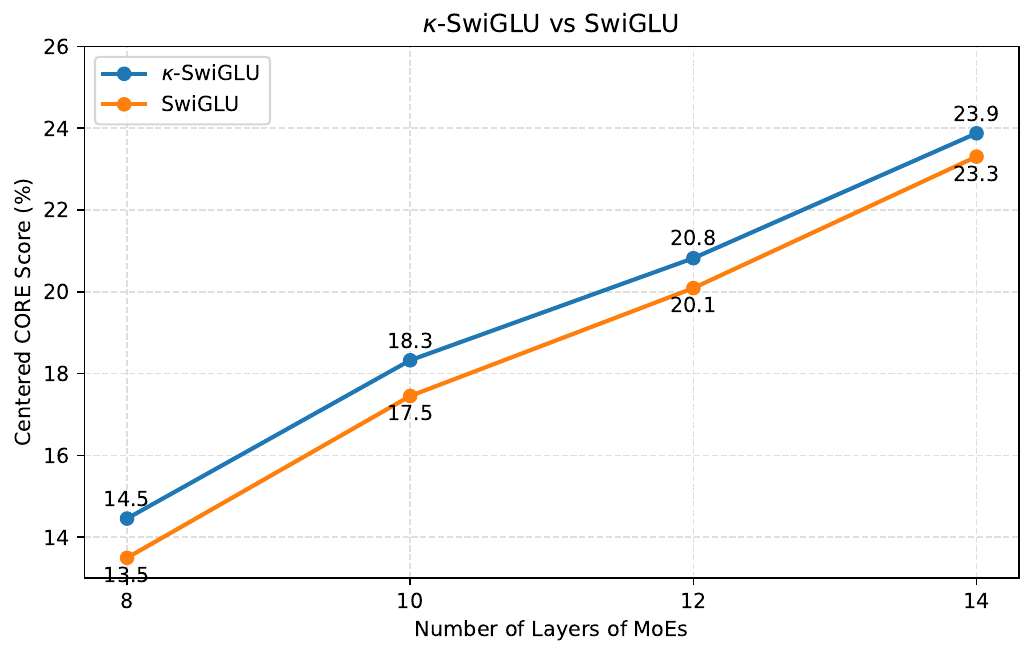}
  \caption{Performance comparison between $\kappa$-SwiGLU and standard SwiGLU across different layers of standard MoE models. The y-axis reports the centered CORE score, computed as the average score across 22 CORE benchmarks relative to a fixed-answer baseline. $\kappa$-SwiGLU improves over standard SwiGLU at all evaluated standard MoE depths.}

\label{fig:moe-scale}
\end{figure}

\begin{figure}[t]
  \centering
  \includegraphics[width=\columnwidth]{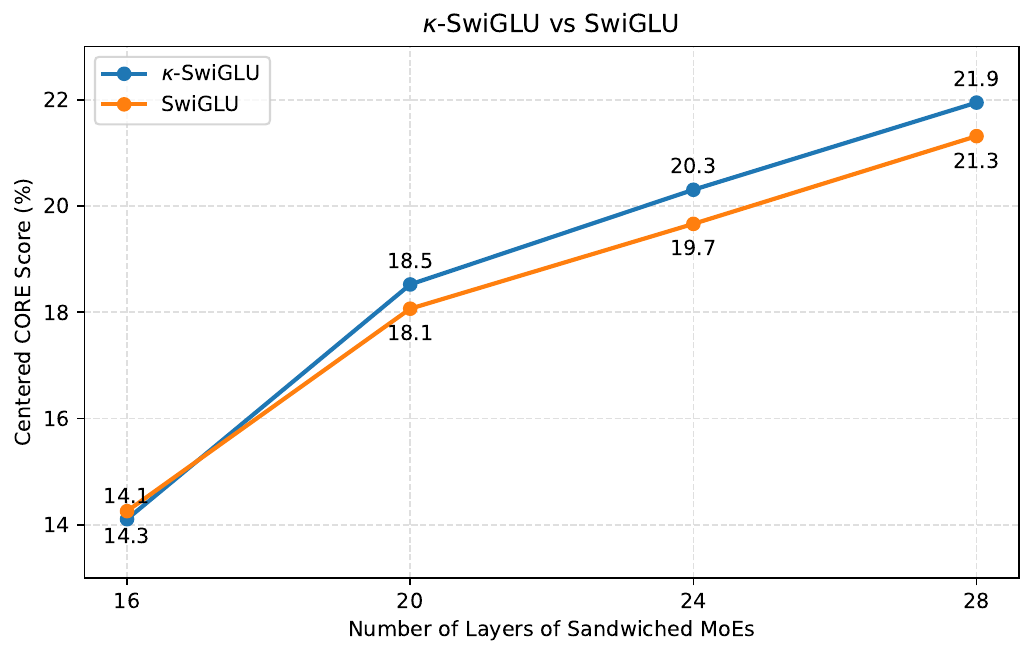}
  \caption{Performance comparison between $\kappa$-SwiGLU and standard SwiGLU across different numbers of total layers in sandwiched MoE models. The y-axis reports the centered CORE score, computed as the average score across 22 CORE benchmarks relative to a fixed-answer baseline. $\kappa$-SwiGLU consistently outperforms standard SwiGLU for models with more than 16 layers, with slightly larger gains at higher layer counts.}

\label{fig:sandwiched-moe-scale}
\end{figure}

\subsection{Main Results}

\begin{table}[t]
\centering
\small
\begin{tabular}{lccc}
\toprule
Model & SwiGLU & $\kappa$-SwiGLU & $\Delta$ \\
\midrule
MoE-8L  & $13.5 \pm 1.0$ & $\mathbf{14.5 \pm 0.4}$ & $+1.0$ \\
MoE-10L & $17.5 \pm 1.2$ & $\mathbf{18.3 \pm 0.9}$ & $+0.9$ \\
MoE-12L & $20.1 \pm 1.0$ & $\mathbf{20.8 \pm 0.2}$ & $+0.7$ \\
MoE-14L & $23.3 \pm 0.3$ & $\mathbf{23.9 \pm 0.6}$ & $+0.6$ \\
\midrule
Sandwich-16L & $\mathbf{14.3 \pm 1.0}$ & $14.1 \pm 0.4$ & $-0.2$ \\
Sandwich-20L & $18.1 \pm 0.3$ & $\mathbf{18.5 \pm 0.7}$ & $+0.5$ \\
Sandwich-24L & $19.7 \pm 0.7$ & $\mathbf{20.3 \pm 1.3}$ & $+0.6$ \\
Sandwich-28L & $21.3 \pm 1.1$ & $\mathbf{21.9 \pm 1.0}$ & $+0.6$ \\
\bottomrule
\end{tabular}
\caption{Centered CORE scores for standard SwiGLU and $\kappa$-SwiGLU. To account for randomness in experimental results, each score is averaged over three independent runs with random seeds 24, 26, and 28; standard deviations are reported after $\pm$. $\Delta$ denotes the difference between the mean scores of $\kappa$-SwiGLU and standard SwiGLU in percentage points.}
\label{tab:core-scores}
\end{table}

Figures~\ref{fig:moe-scale} and~\ref{fig:sandwiched-moe-scale} compare the performance of $\kappa$-SwiGLU and standard SwiGLU across model depths for standard and sandwiched MoE architectures, respectively. Across the evaluated settings, $\kappa$-SwiGLU improves the mean centered CORE score in 7 out of 8 configurations.

In the standard MoE setting, $\kappa$-SwiGLU improves the centered CORE score by approximately 0.6--1.0 percentage points, with slightly larger gains at shallower depths. One possible explanation is that shallower models use larger expert pools, whereas deeper standard MoE models reduce the number of candidate experts from 64 to 16 to fit within the memory budget. This makes the deeper models effectively denser and may reduce the relative benefit of confidence-aware expert gating.

In the sandwiched MoE setting, $\kappa$-SwiGLU yields gains of approximately 0.4--0.6 percentage points for models deeper than 16 layers, with slightly larger gains at greater depths. 

Although the improvement in each individual setting is modest relative to run-to-run variation, $\kappa$-SwiGLU improves the mean centered CORE score in 7 out of 8 settings. This cross-setting consistency suggests that confidence-aware gate sharpness provides a robust positive trend across MoE architectures and model depths, while introducing only negligible additional parameters and small computational overhead.

A full breakdown over the 22 CORE benchmarks is provided in Appendix~\ref{app:core-breakdown}. Across the 22 benchmarks, $\kappa$-SwiGLU improves or matches the baseline on the majority of tasks, suggesting that the gains are not driven by a single benchmark.

In addition, Appendix~\ref{app:kappa-analysis} analyzes the learned $\kappa$ parameters in detail. We show that after the warm-up phase, $\kappa$ rapidly diverges from the standard fixed value of 1, with some gate units becoming substantially sharper and others substantially smoother. Over training, these values gradually return toward a more moderate range while remaining separated from 1, indicating that $\kappa$-SwiGLU learns a nontrivial and persistent modulation of gate sharpness. We further analyze the learned scale and bias parameters, $\alpha$ and $b$, and find that the router-logit-dependent scale term $\alpha_{e,j}s_e(x)$ contributes more strongly than the offset term $b_{e,j}$, supporting the importance of routing confidence in driving the learned sharpness modulation.

\subsection{Ablation Studies}

\begin{table}[t]
\centering
\small
\begin{tabular}{lccc}
\toprule
Method & MoE-8L & MoE-10L & $\Delta$ Avg. \\
\midrule
SwiGLU                         & $13.5 \pm 1.0$ & $17.5 \pm 1.2$ & $-0.9$ \\
$\kappa$-SwiGLU$_{-\alpha}$    & $13.4 \pm 0.6$ & $17.8 \pm 1.0$ & $-0.8$ \\
$\kappa$-SwiGLU$_{-b}$         & $13.9 \pm 0.4$ & $\mathbf{18.5 \pm 0.3}$ & $-0.2$ \\
$\kappa$-SwiGLU                & $\mathbf{14.5 \pm 0.4}$ & $18.3 \pm 0.9$ & $0.0$ \\
\bottomrule
\end{tabular}
\caption{Ablation study of $\kappa$-SwiGLU components on MoE-8L and MoE-10L. $\kappa$-SwiGLU$_{-\alpha}$ removes the router-logit-dependent scale term, $\kappa$-SwiGLU$_{-b}$ removes the offset term, and $\kappa$-SwiGLU denotes the full method. Centered CORE scores are reported.}
\label{tab:kappa-ablation}
\end{table}

We perform ablation studies to understand the contributions of the two components in the $\kappa$ parameterization in Eq.~\eqref{eq:kappa}: the router-logit-dependent scale term $\alpha_{e,j} \cdot s_e(x)$ and the offset term $b_{e,j}$. We compare three variants: $\kappa$-SwiGLU$_{-\alpha}$, which removes the scale term by setting $\alpha_{e,j}=0$; $\kappa$-SwiGLU$_{-b}$, which removes the offset term by setting $b_{e,j}=0$; and the full $\kappa$-SwiGLU method.

As shown in Table~\ref{tab:kappa-ablation}, removing the scale term $\alpha_{e,j} \cdot s_e(x)$ consistently hurts performance. In contrast, removing the offset term $b_{e,j}$ has a smaller impact. This suggests that the confidence-dependent scale term accounts for most of the benefit, while the offset mainly provides additional flexibility.

\subsection{Computational Overhead of $\kappa$-SwiGLU}

\begin{table}[t]
\centering
\small
\begin{tabular}{lccc}
\toprule
Method & \makecell{Active Params\\(M)} & \makecell{Training\\TPS} & \makecell{Inference\\TPS} \\
\midrule
SwiGLU          & 1,097 & 153,200 & 24,600 \\
$\kappa$-SwiGLU & 1,098 & 142,500 & 23,729 \\
\midrule
$\Delta$        & $+0.02\%$ & $-7.0\%$ & $-3.5\%$ \\
\bottomrule
\end{tabular}
\caption{Computational overhead of $\kappa$-SwiGLU compared with standard SwiGLU on the largest MoE-14L model. TPS denotes tokens per second.}
\label{tab:compute-overhead}
\end{table}

Table~\ref{tab:compute-overhead} compares the computational overhead of $\kappa$-SwiGLU with standard SwiGLU on the MoE-14L model, measured by active parameter count and tokens per second (TPS) during training and inference. $\kappa$-SwiGLU introduces only $0.02\%$ additional active parameters. Its inference throughput differs from the standard SwiGLU baseline by only $4.0\%$, indicating that $\kappa$-SwiGLU achieves performance gains with a small computational overhead.

\section{Conclusion}

In this work, we propose $\kappa$-SwiGLU, a confidence-aware variant of SwiGLU for Mixture-of-Experts (MoE) models that dynamically adjusts expert gate sharpness based on token-level routing confidence. By explicitly coupling router logits with expert gate sharpness, $\kappa$-SwiGLU allows each expert to adapt its activation selectivity according to routing confidence, providing a more flexible and expressive gating mechanism. Experiments on FineWeb-Edu show that $\kappa$-SwiGLU improves mean CORE performance across a range of MoE architectures and model depths, while introducing only negligible additional parameters and a small computational overhead. Future work could explore alternative parameterizations of confidence-aware gate modulation, as well as applications of $\kappa$-SwiGLU to MoE models beyond language modeling.

\section*{Limitations}

This work evaluates $\kappa$-SwiGLU on relatively small-scale MoE language models trained on FineWeb-Edu. Although our experiments cover multiple model depths and both standard and sandwiched MoE architectures, the largest models remain much smaller than frontier-scale MoE systems due to limited computational resources. It remains to be verified whether the same trends hold at substantially larger parameter counts, longer training schedules, and larger-scale pretraining corpora.

Our evaluation is primarily based on pretrained model performance measured by CORE. While CORE covers a diverse set of benchmarks, it does not fully capture downstream behavior after instruction tuning, long-context use, reasoning-heavy evaluation, or deployment-oriented metrics. Broader evaluation is needed to better characterize where confidence-aware gate sharpness is most beneficial.

The proposed method introduces only a small number of additional parameters, but it incurs a small computational overhead of approximately $4$--$7\%$ due to the extra elementwise operations needed to compute token-dependent sharpness coefficients. Further kernel-level optimization may reduce this overhead to a negligible level.

Finally, our method parameterizes the sharpness coefficient using a simple affine transformation of router logits followed by a bounded mapping. Other confidence signals, parameterizations, initialization strategies, or regularization schemes may lead to different trade-offs between stability, expressivity, and performance. We leave a more systematic exploration of these design choices, as well as applications beyond language modeling, to future work.

\bibliography{moe}

\clearpage

\appendix
\section{Breakdown of CORE Results}
\label{app:core-breakdown}

The CORE benchmark for pretrained model evaluation consists of 22 datasets, spanning a diverse set of tasks including textbook knowledge, commonsense reasoning, and language modeling. To provide a more fine-grained view of model behavior beyond the aggregate CORE score, Tables~\ref{tab:task-centered-moe} and~\ref{tab:task-centered-sandwich} report per-task performance across all evaluated configurations. 

We note that BoolQ contributes noticeably to the improvement in several settings, especially for sandwiched MoE models. To verify that the observed gains are not solely driven by this benchmark, we also report Centered CORE without BoolQ. As shown in Tables~\ref{tab:task-centered-moe} and~\ref{tab:task-centered-sandwich}, $\kappa$-SwiGLU still improves CORE (no BoolQ) in most settings, although the gains are smaller than in the full CORE average. This suggests that BoolQ amplifies the aggregate improvement, but the benefit of confidence-aware gate sharpness is not entirely explained by a single benchmark.

\begin{table*}[t]
\centering
\scriptsize
\resizebox{\textwidth}{!}{%
\begin{tabular}{lrrrrrrrr}
\toprule
Task 
& \makecell{MoE-8L\\Base} 
& \makecell{MoE-8L\\$\kappa$} 
& \makecell{MoE-10L\\Base} 
& \makecell{MoE-10L\\$\kappa$} 
& \makecell{MoE-12L\\Base} 
& \makecell{MoE-12L\\$\kappa$} 
& \makecell{MoE-14L\\Base} 
& \makecell{MoE-14L\\$\kappa$} \\
\midrule
HellaSwag (0-shot) & \textbf{17.45} & 17.39 & \textbf{24.20} & 24.03 & 28.39 & \textbf{28.44} & \textbf{33.44} & 33.32 \\
Jeopardy & \textbf{1.57} & 1.21 & \textbf{4.98} & 3.94 & 8.44 & \textbf{9.54} & 14.00 & \textbf{14.23} \\
BBH QA Wikidata & 19.81 & \textbf{24.20} & 39.23 & \textbf{41.01} & \textbf{45.11} & 44.12 & \textbf{50.44} & 46.97 \\
ARC-Easy & 38.05 & \textbf{38.14} & \textbf{46.56} & 45.88 & 49.61 & \textbf{49.83} & \textbf{53.83} & 52.97 \\
ARC-Challenge & \textbf{4.85} & 4.59 & 7.58 & \textbf{9.14} & \textbf{12.51} & 12.06 & \textbf{16.46} & 14.94 \\
COPA & 34.67 & \textbf{36.00} & 30.00 & \textbf{32.67} & 28.00 & \textbf{31.33} & 36.67 & \textbf{39.33} \\
CommonsenseQA & 11.34 & \textbf{13.46} & 9.06 & \textbf{13.97} & 2.85 & \textbf{6.98} & 7.32 & \textbf{11.86} \\
PIQA & \textbf{25.64} & 24.92 & 29.38 & \textbf{31.77} & 33.80 & \textbf{35.73} & \textbf{38.77} & 36.53 \\
OpenBookQA & \textbf{9.07} & 8.80 & 10.76 & \textbf{11.73} & 12.18 & \textbf{13.78} & \textbf{16.62} & 13.60 \\
LAMBADA & \textbf{30.22} & 29.48 & \textbf{35.67} & 34.83 & \textbf{37.55} & 36.91 & \textbf{42.03} & 40.04 \\
HellaSwag & \textbf{17.65} & 17.36 & 24.43 & \textbf{24.49} & \textbf{28.68} & 28.66 & \textbf{34.12} & 33.96 \\
Winograd & 15.51 & \textbf{19.90} & 24.30 & \textbf{26.98} & 30.16 & \textbf{30.40} & 34.31 & \textbf{37.48} \\
WinoGrande & \textbf{2.34} & \textbf{2.34} & 4.66 & \textbf{5.29} & 5.97 & \textbf{6.50} & \textbf{9.08} & 7.23 \\
BBH Dyck & \textbf{8.63} & 7.00 & \textbf{8.80} & 5.53 & 12.67 & \textbf{13.17} & 12.00 & \textbf{13.43} \\
LSAT-AR & 5.80 & \textbf{8.88} & 5.43 & \textbf{7.07} & \textbf{6.70} & 5.80 & 4.35 & \textbf{10.87} \\
BBH CS Algorithms & \textbf{39.42} & 38.91 & \textbf{40.71} & 40.48 & 39.19 & \textbf{41.59} & 37.10 & \textbf{37.53} \\
BBH Operators & 12.54 & \textbf{13.33} & 16.51 & \textbf{17.30} & 15.71 & \textbf{15.87} & \textbf{19.05} & 18.57 \\
BBH Repeat Copy & \textbf{2.08} & \textbf{2.08} & \textbf{2.08} & \textbf{2.08} & \textbf{3.12} & \textbf{3.12} & \textbf{1.04} & \textbf{1.04} \\
SQuAD & \textbf{10.39} & 10.32 & 17.86 & \textbf{18.57} & 22.60 & \textbf{22.95} & \textbf{29.01} & 28.06 \\
CoQA & \textbf{12.63} & 12.03 & 16.30 & \textbf{16.37} & 18.66 & \textbf{18.71} & \textbf{21.59} & 21.53 \\
BoolQ & -40.54 & \textbf{-29.92} & -32.60 & \textbf{-28.09} & -17.63 & \textbf{-15.32} & -16.29 & \textbf{-5.67} \\
BBH Lang ID & \textbf{17.76} & 17.60 & 18.06 & \textbf{18.14} & 17.73 & \textbf{17.94} & \textbf{17.87} & 17.47 \\
\midrule
Centered CORE & 13.49 & \textbf{14.46} & 17.45 & \textbf{18.33} & 20.09 & \textbf{20.82} & 23.31 & \textbf{23.88} \\
Centered CORE (no BoolQ) & 16.07 & \textbf{16.57} & 19.84 & \textbf{20.54} & 21.89 & \textbf{22.54} & 25.19 & \textbf{25.28} \\
\bottomrule
\end{tabular}%
}
\caption{Per-task performance for MoE configurations. Task rows report centered accuracy percentages. Centered CORE is computed as the mean of the 22 centered accuracy values. We include Centered CORE (no BoolQ) to assess whether aggregate improvements are driven by a single benchmark. The better score within each matched Base/$\kappa$ pair is bolded.}\label{tab:task-centered-moe}
\end{table*}

\begin{table*}[t]
\centering
\scriptsize
\resizebox{\textwidth}{!}{%
\begin{tabular}{lrrrrrrrr}
\toprule
Task & \makecell{Sandwich-16L\\Base} & \makecell{Sandwich-16L\\$\kappa$} & \makecell{Sandwich-20L\\Base} & \makecell{Sandwich-20L\\$\kappa$} & \makecell{Sandwich-24L\\Base} & \makecell{Sandwich-24L\\$\kappa$} & \makecell{Sandwich-28L\\Base} & \makecell{Sandwich-28L\\$\kappa$} \\
\midrule
HellaSwag (0-shot) & \textbf{16.71} & 16.55 & 22.94 & \textbf{23.12} & \textbf{26.91} & 26.71 & \textbf{31.67} & 31.56 \\
Jeopardy & \textbf{1.40} & 1.24 & \textbf{5.76} & 4.90 & 4.11 & \textbf{6.22} & \textbf{10.60} & 9.94 \\
BBH QA Wikidata & 29.17 & \textbf{29.55} & 39.80 & \textbf{40.04} & \textbf{42.75} & 41.59 & 46.29 & \textbf{47.16} \\
ARC-Easy & 39.60 & \textbf{39.69} & 47.59 & \textbf{47.62} & \textbf{49.78} & 49.29 & \textbf{52.84} & 52.56 \\
ARC-Challenge & 4.55 & \textbf{4.66} & \textbf{8.68} & 8.12 & \textbf{12.02} & 10.58 & 13.84 & \textbf{14.87} \\
COPA & 23.33 & \textbf{29.33} & 28.67 & \textbf{34.67} & \textbf{27.33} & 24.00 & \textbf{28.67} & 26.00 \\
CommonsenseQA & 11.96 & \textbf{12.37} & 11.96 & \textbf{13.94} & 11.45 & \textbf{12.09} & 7.32 & \textbf{9.75} \\
PIQA & \textbf{29.13} & 29.02 & 34.60 & \textbf{35.11} & 34.75 & \textbf{37.18} & 39.93 & \textbf{40.88} \\
OpenBookQA & 8.71 & \textbf{9.60} & \textbf{13.16} & 12.53 & 14.40 & \textbf{14.76} & 15.56 & \textbf{16.27} \\
LAMBADA & \textbf{30.53} & 29.41 & \textbf{33.88} & 33.56 & 35.34 & \textbf{36.73} & 39.12 & \textbf{39.43} \\
HellaSwag & \textbf{16.81} & 16.79 & \textbf{23.16} & 23.09 & \textbf{27.21} & 26.98 & \textbf{32.16} & 31.98 \\
Winograd & 12.82 & \textbf{15.75} & \textbf{24.30} & \textbf{24.30} & \textbf{27.72} & 26.25 & \textbf{37.00} & 35.33 \\
WinoGrande & 2.13 & \textbf{3.87} & 6.08 & \textbf{6.71} & 4.66 & \textbf{7.97} & 7.23 & \textbf{8.50} \\
BBH Dyck & 10.70 & \textbf{10.73} & \textbf{12.43} & 12.03 & 13.03 & \textbf{14.73} & 11.87 & \textbf{12.03} \\
LSAT-AR & 6.34 & \textbf{7.25} & \textbf{8.51} & 5.80 & \textbf{7.61} & \textbf{7.61} & \textbf{7.25} & 4.71 \\
BBH CS Algorithms & \textbf{41.69} & 41.21 & \textbf{42.07} & 41.57 & 40.58 & \textbf{40.83} & \textbf{40.15} & 39.52 \\
BBH Operators & 11.43 & \textbf{11.75} & 15.71 & \textbf{15.87} & \textbf{15.08} & 14.92 & 17.46 & \textbf{18.41} \\
BBH Repeat Copy & \textbf{1.04} & \textbf{1.04} & 1.04 & \textbf{2.08} & 2.08 & \textbf{4.17} & \textbf{0.00} & \textbf{0.00} \\
SQuAD & 11.82 & \textbf{13.11} & 17.51 & \textbf{17.92} & \textbf{22.50} & 21.55 & 26.67 & \textbf{28.53} \\
CoQA & \textbf{13.73} & 13.67 & \textbf{15.74} & 14.77 & \textbf{18.34} & 17.96 & \textbf{20.66} & 20.61 \\
BoolQ & \textbf{-27.77} & -44.37 & -33.54 & \textbf{-28.39} & -23.13 & \textbf{-13.50} & -35.52 & \textbf{-23.01} \\
BBH Lang ID & 17.88 & \textbf{18.14} & 17.44 & \textbf{18.17} & 18.09 & \textbf{18.10} & \textbf{18.15} & 17.77 \\
\midrule
Centered CORE & \textbf{14.26} & 14.11 & 18.07 & \textbf{18.52} & 19.66 & \textbf{20.31} & 21.31 & \textbf{21.94} \\
Centered CORE (no BoolQ) & 16.26 & \textbf{16.89} & 20.52 & \textbf{20.76} & 21.70 & \textbf{21.91} & 24.02 & \textbf{24.09} \\
\bottomrule
\end{tabular}%
}
\caption{Per-task performance for Sandwiched-MoE configurations. Task rows report centered accuracy percentages. Centered CORE is computed as the mean of the 22 centered accuracy values. We include Centered CORE (no BoolQ) to assess whether aggregate improvements are driven by a single benchmark. The better score within each matched Base/$\kappa$ pair is bolded.}\label{tab:task-centered-sandwich}
\label{tab:task-centered-exp128}
\end{table*}

\section{Empirical Analysis of $\kappa$-SwiGLU}
\label{app:kappa-analysis}

\begin{figure}[t]
  \centering
  \includegraphics[width=\columnwidth]{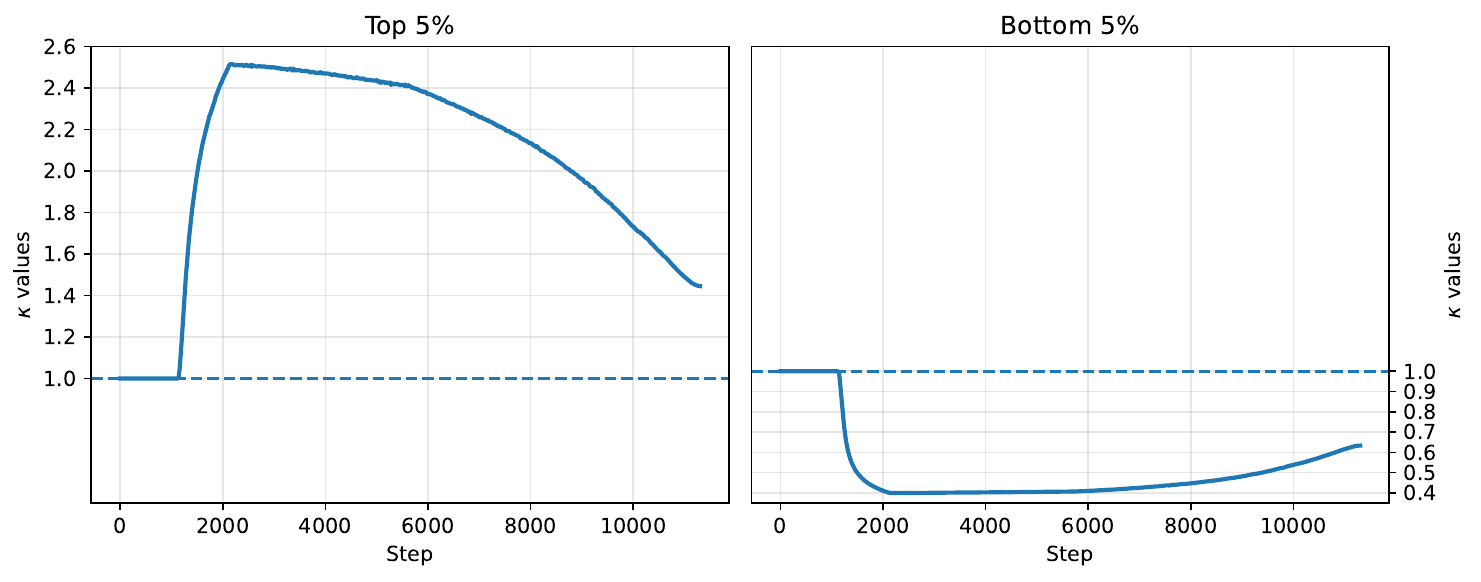}  
\caption{The mean of the top and bottom $5\%$ of the learned $\kappa$ values in the 9th layer of a 12-layer MoE. 
During the first 1,100 training iterations, the $\kappa$ values are frozen at 1, corresponding to the standard SiLU gate. 
Afterward, they rapidly diverge: the top $5\%$ increase to around $2.5$, while the bottom $5\%$ decrease to around $0.4$. 
This indicates that $\kappa$-SwiGLU initially learns both sharper, more selective gates and smoother, more broadly active gates. 
In the second half of training, the values gradually move back toward 1, suggesting that the model later adopts a more moderate modulation of gate sharpness.
}\label{fig:kappa-values}
\end{figure}

\begin{figure}[t]
  \centering
  \includegraphics[width=\columnwidth]{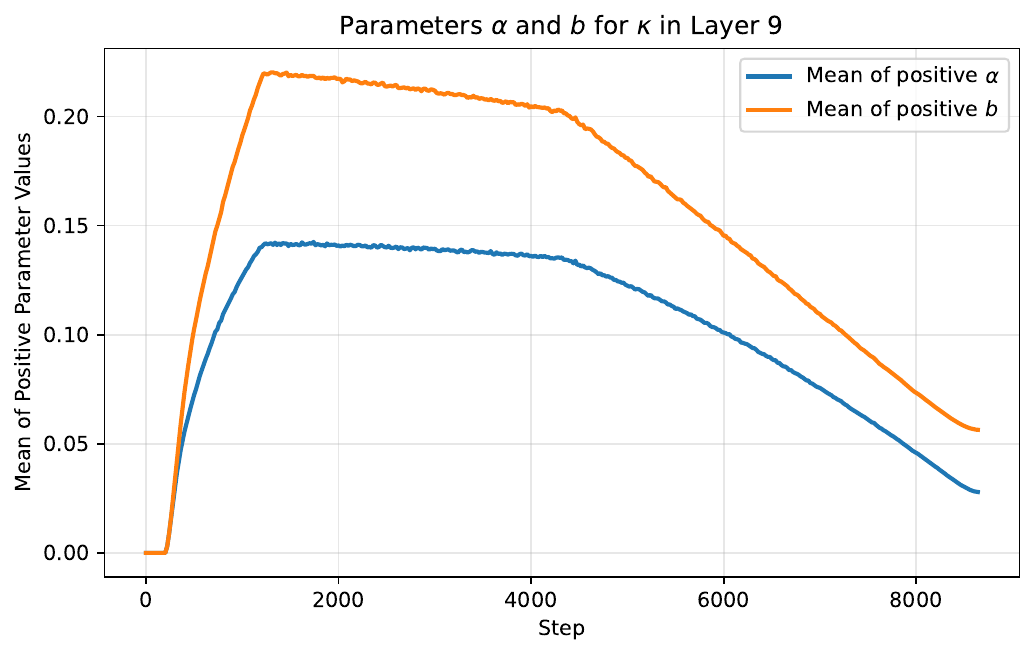}
\caption{Mean of the positive subsets of the learned scale and bias parameters, $\alpha$ and $b$, in the $\kappa$ parameterization in Eq.~\eqref{eq:kappa} for the 9th layer of a 12-layer MoE. Both $\alpha$ and $b$ follow trends similar to the resulting sharpness coefficient $\kappa$. The negative subsets of $\alpha$ and $b$ are approximately symmetric.}
\label{fig:kappa-scale-bias}
\end{figure}

Figure~\ref{fig:kappa-values} shows the training dynamics of the top and bottom $5\%$ of the learned $\kappa$ values in a representative layer of a 12-layer MoE, with similar behavior observed across all MoE layers.

During the initial warm-up phase, the $\kappa$ values remain fixed at 1, reducing $\kappa$-SwiGLU to the standard SiLU gate. 
Once unfrozen, the two groups rapidly diverge: tokens with the largest learned $\kappa$ values move toward much sharper gates, while those with the smallest learned $\kappa$ values move toward smoother gates. 
Interestingly, this separation is strongest shortly after unfreezing, with the top $5\%$ reaching around $2.5$ and the bottom $5\%$ dropping to around $0.4$. As training proceeds, both groups gradually move back toward 1, suggesting that the model initially explores a wide range of gate sharpness but later settles on a more moderate modulation. 
However, by the end of training, both groups remain substantially separated from 1, indicating that the learned $\kappa$ values continue to have a non-negligible effect on the gating behavior.

This behavior indicates that $\kappa$-SwiGLU introduces a flexible, input-dependent adjustment of gate selectivity rather than simply making all gates uniformly sharper or smoother.

Since the $\kappa$ parameterization in Eq.~\eqref{eq:kappa} contains two learnable parameters, $\alpha$ and $b$, we further track their training dynamics to understand their relative contributions to the learned $\kappa$ values. We observe that the positive and negative values are approximately symmetric; therefore, we report only the mean of the positive values, with the negative values following the same trend up to a sign flip. As shown in Figure~\ref{fig:kappa-scale-bias}, both parameters exhibit trends similar to the resulting $\kappa$: they increase sharply during the early phase of training, then gradually decay toward zero while remaining around $0.05$ by the end of training. Because $\alpha$ is multiplied by the router logit $s_e(x)$, which empirically usually falls in the range $[2,4]$, the term $\alpha \cdot s_e(x)$ contributes substantially more to $\kappa$ than the bias term $b_{e,j}$. For example, using a typical router logit value of $s_e(x)=2.5$, the scale term in the middle of training contributes approximately $0.134 \times 2.5 = 0.335$ to the affine input of $\phi$. This is about $1.675\times$ the bias contribution $b=0.2$, indicating that the confidence-dependent scale term dominates the learned sharpness modulation.

\section{Scaling of Computational Budgets}
\label{app:scaling}
Since MoE models contain substantially more total parameters than dense models, we use a token-to-parameter ratio of 5 for all models. Following the common observation that not all MoE parameters are activated for each token, we estimate the effective parameter count using a square-root scaling rule
\begin{align}
N_{\mathrm{eff}}
=
N_{\mathrm{shared}}
+
N_{\mathrm{attn}}^{\mathrm{MoE}}
+
N_{\mathrm{MLP}}^{\mathrm{all\ experts}}
\sqrt{\frac{k}{E}}.
\end{align}
where $k=2$ is the top-$k$ routing value, and $E$ is the total number of experts. This rule discounts the expert MLP parameters by $\sqrt{k/E}$ rather than the raw active fraction $k/E$, assigning a larger effective parameter count to the expert pool. This is consistent with the intuition that sparse routing makes the effective capacity of MoE layers grow sublinearly with the total expert parameter count, while still exceeding the capacity implied by the active parameter count alone.

\end{document}